**PaveSAM – Segment Anything for Pavement Distress**

**Neema Jakisa Owor**
PhD Student
University of Missouri-Columbia
Civil and Environmental Engineering Department
Email: neemajakisa@gmail.com , nodyv@missouri.edu

**Yaw Adu-Gyamfi**
Associate Professor
University of Missouri-Columbia
Civil and Environmental Engineering Department
Email: adugyamfiy@missouri.edu

**Armstrong Aboah**
Assistant Professor
North Dakota State University
Civil and Environmental Engineering
Email armstrong.aboah@ndsu.edu

**Mark Amo-Boateng**
Post-doctoral Fellow
University of Missouri-Columbia
Civil and Environmental Engineering Department
Email: marbz@missouri.edu

**ABSTRACT**

Automated pavement monitoring using computer vision can analyze pavement conditions more efficiently and accurately than manual methods. Accurate segmentation is essential for quantifying the severity and extent of pavement defects and consequently, the overall condition index used for prioritizing rehabilitation and maintenance activities. Deep learning-based segmentation models are however, often supervised and require pixel-level annotations, which can be costly and time-consuming. While the recent evolution of zero-shot segmentation models can generate pixel-wise labels for unseen classes without any training data, they struggle with irregularities of cracks and textured pavement backgrounds. This research proposes a zero-shot segmentation model, PaveSAM, that can segment pavement distresses using bounding box prompts. By retraining SAM's mask decoder with just 180 images, pavement distress segmentation is revolutionized, enabling efficient distress segmentation using bounding box prompts, a capability not found in current segmentation models. This not only drastically reduces labeling efforts and costs but also showcases our model's high performance with minimal input, establishing the pioneering use of SAM in pavement distress segmentation. Furthermore, researchers can use existing open-source pavement distress images annotated with bounding boxes to create segmentation masks, which increases the availability and diversity of segmentation pavement distress datasets.

**Keywords:** Pavement Maintenance, Zero-shot segmentation, Deep Learning, Pavement Distress/Crack Segmentation, SAM

**INTRODUCTION**

Automated pavement monitoring (C. Zhang et al., 2022),(Ghosh & Smadi, 2021),(Li et al., 2021), (Tello-Cifuentes et al., 2024) leveraging computer vision saves time, and ensures safety (Majidifard et al., 2020). Earlier studies on automatic detection used bounding box detection, which draws a box around the distress area. However, this method only provides the location and type of the distress, and does not give enough information for pavement condition analysis. For example, the width and extent of the distress, which are important factors to calculate the Pavement Condition Index of the road (Majidifard et al., 2020), (Owor et al., 2023) and determining the appropriate interventions, cannot be obtained from the bounding box. Therefore, segmentation emerged as a more refined method for pavement distress detection. Segmentation classifies every pixel in an image, and thus gives a more detail about the pavement distress. This can help to evaluate the pavement condition and plan the suitable repairs(Owor et al., 2023). The segmented images can reveal the type, severity, and extent of pavement distresses based on their shape, pattern, width, and area. Accurate segmentation of pavement distresses is essential to avoid unnecessary or excessive maintenance costs(Kheradmandi & Mehranfar, 2022).

Segmentation methods can be either rule-based or deep-learning based. Rule-based methods use predefined features such as thresholding or edge detection to segment the pixels, while deep-learning methods use data-driven features learned by models such as Artificial Neural Networks and Convolutional Neural Networks (Kheradmandi & Mehranfar, 2022). Thresholding methods separate the image into foreground and background pixels by using a threshold value, which can be global or local (Chua & Xu, 1994), (Gonzalez et al., 2009). Some examples of thresholding techniques are Optimal Thresholding (Groenewald et al., 1993), Quadtree (Wu et al., 1982),

Dynamic Edge-Based (Fu & Mui, 1981), and Between Class Variance (Otsu, 1979). Edge detection methods segment the image based on discontinuities in color, light, or texture, which indicate the edges. Filters are often used to detect the edges, such as Shi-Tomasi Algorithm, Sobel Filter, Canny Operator (Tsai et al., 2010), Robert Operator, and Prewitt Operator (Guan et al., 2023). Wang et al. use multi-scale and local optimum thresholding based on crack density, showing more effective and robust results compared to traditional thresholding algorithms (S. Wang & Tang, 2011). (Ouyang & Wang, 2012) applied Otsu's thresholding segmentation algorithm and beamlet transform to extract linear crack features from binary images at various scales and thresholds. (Nnolim, 2020) proposes crack segmentation using wavelet coefficient adjustment, nonlinear filter pre-processing, saturation channel extraction, adaptive threshold-based edge detection, and fuzzy clustering-based area classification. (Sun et al., 2016) developed a pavement crack detection method with uniform background establishment and a weighted neighborhood pixels approach, demonstrating accuracy, speed, and robustness through local thresholding and shape filtering. (Zuo et al., 2008) present a segmentation method based on fractal theory that effectively identifies pavement crack edges. (Nguyen et al., 2011) introduced a crack detection method combining brightness and connectivity for segmentation in road pavement images, achieving accurate detection of cracks of any shape or orientation without requiring defect texture learning. While rule-based methods deliver accurate segmentation results, they are limited by manually defined features and may struggle with diverse image types. Hence, the dominance of deep learning methods in recent years has emerged as a superior alternative, as they automatically learn features from the data, ensuring adaptability and robust performance across various image datasets.

Deep learning neural networks, such as Convolutional Neural Networks (CNNs), Deep Convolutional Neural Networks (DCNNs), and various pretrained models, such as YOLO (Redmon et al., 2015), UNet (Ronneberger et al., 2015), MobileNetV2 (Howard et al., 2017), DeepLab (Chen et al., 2016), GoogleNet (Szegedy et al., 2014), and SqueezeNet (Iandola et al., 2016) have been deployed for image segmentation. These models can segment pavement distresses accurately. Deep Neural Networks learn the image features automatically and can be classified into three types: CNNs, Recurrent Neural Networks (RNNs), and Generative Adversarial Networks (GANs) (Guan et al., 2023). CNNs use convolutional and pooling layers to reduce and extract the image features, such as YOLO, UNet, and DeepLab. RNNs use recurrent layers to learn sequential data, such as Mask R-CNN. GANs use a generator and a discriminator to produce realistic images and distinguish them from synthetic ones. (Lau et al., 2020) present a U-Net-based network architecture for pavement distress segmentation, leveraging a pretrained ResNet-34 encoder and employing a 'one-cycle' training schedule with cyclical learning rates, achieving high F1 scores on the Crack500 datasets through enhancements like squeeze and excitation modules. (Han et al., 2022) address the challenge of interrupted cracks and background noise misidentification in current semantic segmentation methods by introducing CrackW-Net, a novel pixel-level segmentation network utilizing skip-level round-trip sampling blocks within convolutional neural networks. (Kang & Feng, 2022) addresses the challenge of accurately segmenting road surface cracks on complex backgrounds by utilizing a conditional generative adversarial network (GAN) with a U-net3+ generator and attention module to enhance crack feature extraction and suppress noise. (Y. Zhang et al., 2023) tackles the problems with UNet's ability to segment tiny pavement cracks by proposing ALP-Unet, a U-shaped network incorporating an attention module and Laplacian pyramid to enhance boundary information and improve segmentation accuracy. (Jun et al., 2022) proposed an enhanced U-Net model incorporating atrous convolution and attention mechanisms, leveraging ResNet50 for feature extraction, to enhance crack detail and global context for improved crack location accuracy. These papers showcase the progress in deep-learning segmentation for pavement crack detection, emphasizing the integration of diverse techniques to enhance both accuracy and efficiency. However, these segmentation models based on deep learning methods are mostly supervised models, which means they need pixel-level annotations for the classes they are trained to predict. These pixel-level annotations are very costly and time-consuming, especially for pavement distresses, which requires professional knowledge. Segmentation is a more refined method than bounding box detection, but also more tedious. For example, it takes 78s to segment one pavement distress, while it takes only 10s to annotate it with a bounding box (H. Zhang et al., 2022). Unsupervised learning can reduce the annotation cost, but it does not give good results, as it lacks the position and edge information of the images. A compromise between the two (supervised and unsupervised) is weakly supervised segmentation, which can generate pixel-wise labels based on sparse or coarse labels. Some examples of weakly supervised models are Class Activation Map (CAM), Puzzle CAM, Scribbleup, Bounding box Attribution Map (BBAM). (G. Li et al., 2020 propose a semi-supervised method for pavement crack detection that utilizes adversarial learning and a full convolution discriminator, achieving high detection accuracy of 95.91% with only 50% labeled data. (W. Wang & Su, 2021) propose a network that uses a student-teacher model architecture with EfficientUNet for multi-scale crack feature extraction enhancing robustness through adding noise and achieves a high accuracy with reduced annotation workload, yielding F1 scores of 0.6540 and 0.8321 on using only 60% annotated data. Weakly Supervised Learning U-Net (WSL U-Net) (Tang et al., 2022) uses weakly labelled images(crack and non-crack pixel labels) to reduce annotation costs, , and isolated pixel masks to reduce noise, improving generalization of the model. (X. Liu et al., 2023) implemented a cross-consistency training–based semi-supervised segmentation approach, utilizing modified encoder outputs to ensure consistency between primary and secondary decoder predictions, achieving superior performance compared to models trained on the full dataset with only 60% of the annotated data. These papers demonstrate the effectiveness of semi-supervised learning approaches in reducing the annotation workload and improving the accuracy of pavement crack segmentation.

However, a common limitation across all segmentation models—whether supervised, unsupervised, or weakly supervised—is their inability to effectively segment classes that they were not explicitly trained to recognize(Pourpanah et al., 2022). This lack of transferability and robustness poses a significant challenge when applying these models to diverse datasets. Moreover, these models typically require a large number of annotated images for accurate performance, which can be costly and labor-intensive, relying heavily on expert input. In recent years, zero-shot segmentation models have garnered increased attention due to their ability to generate pixel-wise labels for unseen classes without any prior training data, thus reducing the costs associated with data labeling. These models leverage high-level descriptions of classes to establish connections between new (unseen) and familiar (seen) classes. For instance, ZS3, the pioneering zero-shot segmentation model, employs a generative approach to synthesize pixel-level features for unseen classes based on their semantic representations. By blending these synthetic features with real features from known classes, the model can adapt and segment both familiar and unfamiliar classes (Bucher et al., 2019). More recently, SAM (Kirillov et al., 2023) is most recent zero-shot model represents a cutting-edge zero-shot model capable of segmenting any natural image using input prompts such as points, bounding boxes, or text. Trained on an extensive dataset of one billion masks, SAM demonstrates remarkable versatility. However, SAM's performance is less effective when applied to pavement distresses, given their distinct shape and texture characteristics compared to natural images. To address this challenge specifically for pavement distress analysis, the authors propose PaveSAM (Zero-Shot Segmentation Model), designed to seamlessly integrate into automated pavement management systems. Unlike current segmentation models, PaveSAM requires no prior training on pavement distress datasets, enabling immediate application to different pavement images. Leveraging transferable capabilities derived from SAM's extensive training on diverse datasets, PaveSAM offers broader applicability beyond its original training scope.

SAM was fine-tuned specifically for pavement distress segmentation using a dataset comprising 180 images featuring 5 different types of pavement distresses. The adapted model, named PaveSAM, represents the first endeavor to tailor SAM for this task. PaveSAM demonstrated superior performance compared to SAM and other state-of-the-art models on both our dataset and the publicly available crack500 dataset, achieving zero-shot segmentation of pavement distresses with just 180 images and 100 epochs of fine-tuning. This study makes significant contributions to the field of pavement maintenance by introducing a robust model capable of segmenting any pavement distress with minimal input. Specifically, our contributions are as follows:

- The authors fine-tuned a zero-shot segmentation model, PaveSAM, using bounding box prompts—an innovative and efficient approach not currently supported by existing segmentation models. This approach substantially reduces the labeling effort and segmentation costs.

- PaveSAM achieved high performance with only 180 images for fine-tuning, demonstrating its effectiveness with limited input and reducing the time and resources required for model training.

- Our approach facilitates the utilization of existing open-source pavement distress images annotated with bounding boxes to generate segmentation masks, thereby enhancing the availability and diversity of pavement distress datasets for segmentation tasks.

- This research represents the first attempt to fine-tune SAM, a state-of-the-art zero-shot segmentation model, specifically for pavement distress segmentation, further advancing the capabilities of zero-shot learning in this domain.

## METHODS AND MATERIALS

This section provides an in-depth analysis of the dataset utilized in the research, the steps taken to prepare the data for model training, the specific model employed for training, and the model hyperparameters that were taken into account.

**Dataset**

Our dataset consists of top-down images collected using Laser Crack Measurement System (LCMS). It consisted top-down pavement images, each of size 2011 x1014 pixels. The pavement distress masks were generated by creating polygon-based annotations in Computer Vision Annotation Tool (CVAT). These images contained 5 types of pavement distresses which include: longitudinal crack, transverse crack, alligator crack, block crack, patch. Manholes were also annotated to distinguish them from the background and other pavement distresses, minimizing confusion during training and inferencing. The images were split into 75% (180 images) for training and 25% (60 images) for testing. The number of annotations corresponding to the unique pavement distresses are shown in Table 1.

*Table 1: Number of annotations corresponding to the different pavement distresses.*

| | Distress | Number |
|---|---|---|
| 1. | Transverse | 276 |
| 2. | Longitudinal | 400 |
| 3. | Alligator | 218 |
| 4. | Block | 78 |
| 5. | Patch | 74 |
| 6. | Manhole | 79 |

PaveSAM and other state of the art models were also trained and tested on Crack500 dataset (Yang et al., 2019). The CRACK500 dataset consists of 500 images captured on the main campus of Temple University using cell phones, with each image sized around 2,000 × 1,500 pixels. This dataset includes pixel-level annotated binary images. The dataset is partitioned into 250 training images, 50 validation images, and 200 test images for model development and evaluation.

**SAM architecture**

SAM is a prompt-based segmentation model that can segment any object in any image with minimal user input. SAM consists of three main components: an image encoder, a prompt encoder, and a mask decoder, as illustrated in Figure 1.

The image encoder is a masked autoencoder (MAE) (He et al., 2021) pre-trained vision transformer (ViT) that converts the input high resolution images (e.g. 1024 x1024) images into feature vectors. The image encoder produces a 16× down sampled image embedding (64×64). The prompt encoder is a module that embeds the user-provided prompts into the same feature space as the image encoder. The prompts can be sparse (such as points, text, or bounding boxes) or dense (such as masks), and they are represented by positional encodings and learned embeddings. For points, SAM uses Fourier positional encoding(Tancik et al., 2020) and two learnable tokens for foreground and background. For boxes, SAM uses the point encoding of the two opposite corners. For texts, SAM uses the pre-trained text-encoder in CLIP. The mask decoder is a modified transformer decoder that takes the outputs of the image and prompt encoders and generates segmentation masks of the desired objects. The mask decoder also has a dynamic prediction head and Intersection Over Union(IOU) score regression head. The mask prediction head outputs three masks with four times down sampled masks representing whole object, part, and subpart of the object (Ma et al., 2023).

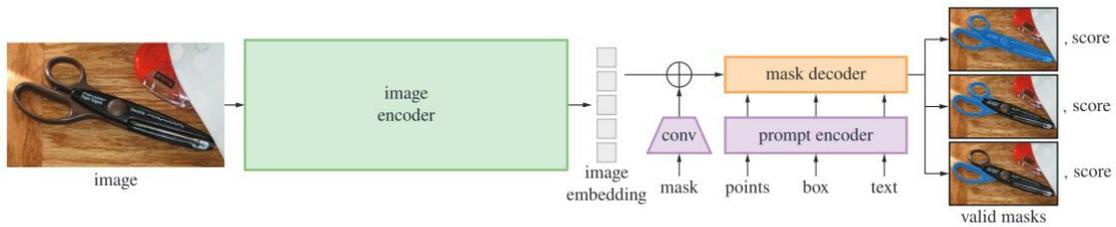

*Figure 1: SAM Architecture (Kirillov et al., 2023)*

**SAM on Pavement Distress Segmentation**

SAM can segment objects in images using different types of prompts: text, points, bounding boxes, or segment everything mode. However, the text-based prompt code is not publicly available yet, so we used the other two prompts to segment pavement distresses. Figure 2 shows the segmentation results from the Segment Anything Demo Website. The segment everything mode does not perform well on pavement distresses, as seen in Figure 2b, 2f, and 2j. This is because pavement distresses are thin and irregular, unlike the natural images that SAM was trained on. The segment everything mode only works well on the manhole, (Figure 2j) which is wider and more regular than pavement distresses. Moreover, the segment everything mode divides the image into sections that are irrelevant for pavement distress segmentation as shown in Figure 2j. For the point prompts, as shown in Figure 2c,2g, and 2k, SAM also fails to segment the linear cracks. In Figure 2g, it segments one of the block cracks, but the segmentation is too wide. The point prompt also works well on the manhole as shown in Figure 2k. Using the bounding box prompt, SAM achieves better segmentation for the pavement distresses than the previous two prompts. However, the segmentation is still too wide and inaccurate for further pavement condition analysis, as seen in Figure 2d, 2h, and 2l. This reflects the difficulty of segmenting pavement distresses with SAM, hence the need to finetune it for Pavement distress segmentation.

| Original Image | Segment Everything prompt | Point prompt | Bounding box prompt |
|---|---|---|---|
| 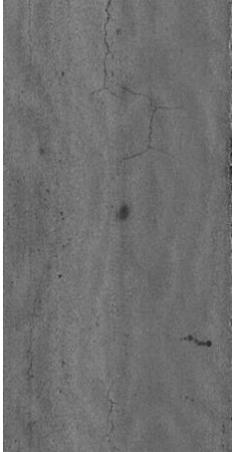 | 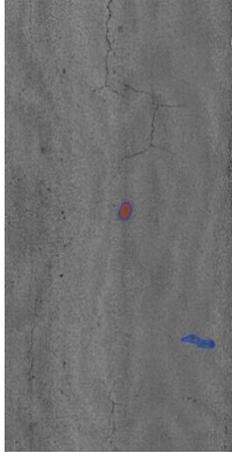 | 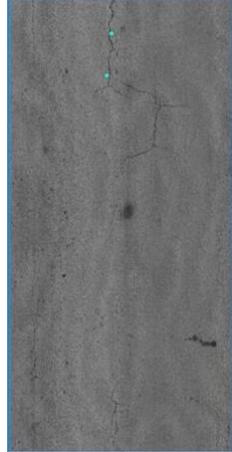 | 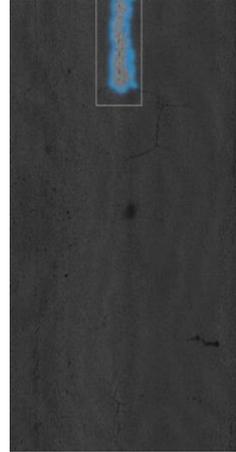 |
| ( a ) | (b) | (c) | (d) |

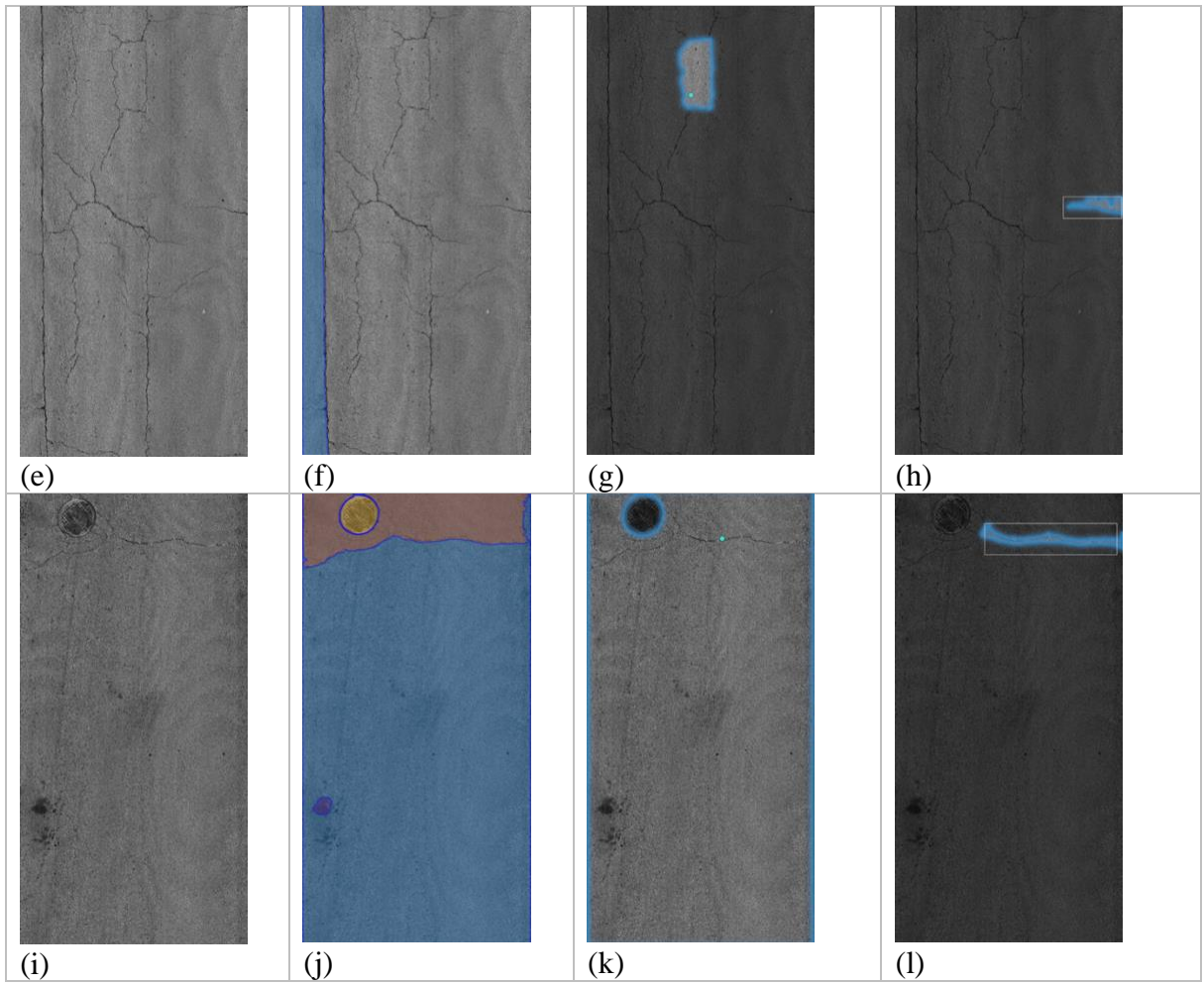

*Figure 2: Segmentation of pavement distresses using different prompts*

**PaveSAM Architecture**

Among the different prompt types, the bounding box prompt produced the best segmentation results for pavement distresses. Also because there are many open source databases that have bounding box annotations and researchers could use these to increase the availability of segmentation data. Therefore, SAM is finetuned with the bounding box prompt to create a pavement distress-specific segmentation model, called PaveSAM. Moreover, researchers could use the open source databases that have bounding box annotations to increase the availability of segmentation data using PaveSAM. The architecture of PaveSAM is shown in Figure 3. PaveSAM has three components: an image encoder, a prompt encoder, and a mask decoder.

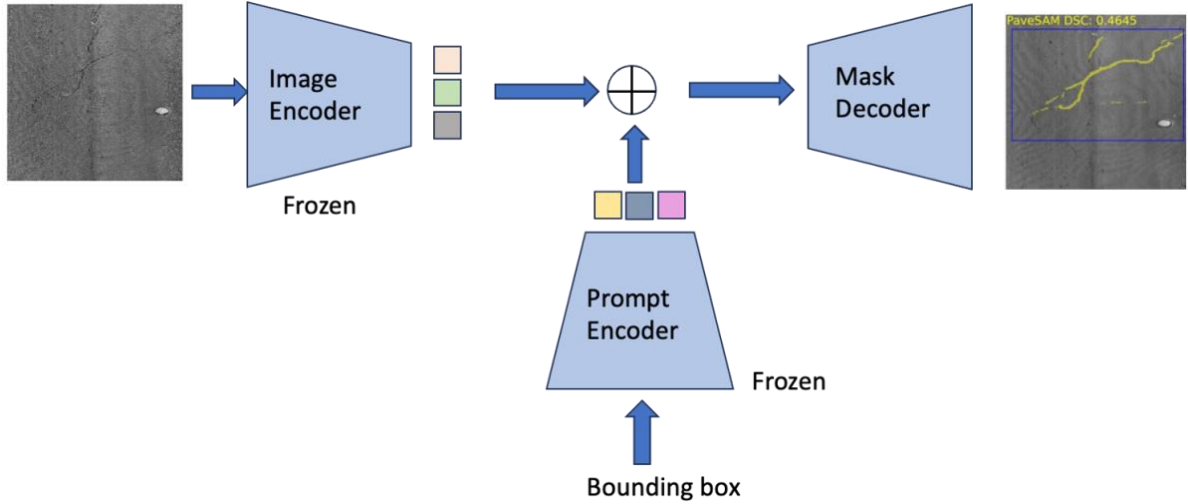

*Figure 3: PaveSAM architecture*

**Image encoder**

The image encoder employed is a pretrained vision transformer (ViT), specifically the Masked AutoEncoder (MAE) pretrained ViT model, to convert pavement images into feature vectors, as illustrated in Figure 3. During training of the MAE (He et al., 2021), random patches within images are masked out, and the model is trained to reconstruct the missing pixels, effectively reconstructing images from masked inputs. Consequently, the encoder can effectively capture essential image components and encode them into relevant latent representation of the images. The encoder was selected based on its capacity to scale and its robust training methodology. The pretrained ViT base model is designed to handle input sizes of 3 x 1024 x 1024. Given that our images are 1014×2011, they are resized and padded on the shorter side to meet the model's input specifications. As the encoder downscales the image by a factor of 16, the output of the image encoder is reduced to 16 x 16 (Kirillov et al., 2023).

**Prompt encoder**

The prompt encoder is a module that embeds the bounding box coordinates into the same feature space as the image encoder. To extract the bounding box of pavement distress using ground truth masks, a python code was used to identify the top-left ($x_{min}$, $y_{min}$) and bottom-right ($x_{max}$, $y_{max}$) coordinates where pixel values exceeded a threshold of 1. These two corner coordinates are then mapped by the encoder into a pair of 256-dimensional embeddings. This embedding pair combines position encoding with learned embeddings to represent the bounding box within the feature space (Kirillov et al., 2023).

**Mask decoder**

The mask decoder is an architecture based on the Transformer model that provides a mask as output. It achieves this by translating image embeddings(from image encoder) and prompt embeddings(from the prompt encoder) to the mask. The decoder incorporates self-attention and cross-attention methods to enhance the embeddings and seamlessly integrate prompt information into the image embeddings. Positional encodings are included to guarantee the preservation of crucial geometric information. The decoder's outputs are up sampled using transposed convolutional layers and further refined by a Multi Layer Perceptron to generate a mask that matches the dimensions of the upscaled picture embeddings(Kirillov et al., 2023).

**Training**

To reduce the computational cost, the image encoder is frozen and the prompt encoder, and only fine-tuned the mask decoder for pavement distress segmentation.

Masks are generated for 180 top-down pavement images using polygon-based annotations on CVAT. These images contained 5 types of pavement distresses which include: longitudinal crack, transverse crack, alligator crack, block crack, patch, manhole.

The model was trained for 100 epochs. The loss was the summation of the dice loss and the cross-entropy loss a robust loss for segmentation tasks. Adam optimizer was used, and a learning rate of 1e-5 was used.

**Loss**

Cross-entropy loss ( see Equation 1) is a commonly used loss function in machine learning for classification problems. It is defined as the difference in probability distributions between two random variables or sets of events(F. Liu & Wang, 2022). The choice of cross entropy loss is motivated by its ability to minimize pixel-level errors by penalizing differences between predicted probability and actual labels, hence favoring more accurate predictions (Yeung et al., 2022).

$$L_{BCE}(y, \hat{p}) = -[ylog\hat{p} + (1 - y)\log(1 - \hat{p})] \qquad (1)$$

The Dice coefficient is a popular metric in the computer vision domain for calculating the similarity of two images. In 2016, it was also developed as a loss function known as Dice Loss. Dice loss (Equation 2) is defined as 1 minus the dice coefficient(F. Liu & Wang, 2022). The benefit of utilizing dice loss is its ability to effectively address the imbalance in pixel count between the background and foreground classes (Yeung et al., 2022).

$$L_{DC}(y, \hat{p}) = 1 - \frac{2y\hat{p}+1}{y + \hat{p}+1} \qquad (2)$$

In circumstances when there is an imbalance in class distribution, the use of cross entropy as a loss function might cause larger objects to be heavily represented, leading to a lower quality segmentation of smaller objects. Hence, the Total loss (Equation 3) is the combination of the Cross entropy loss and the Dice loss, as defined in Equation 3. This combination enhances the diversity and stability of the cross entropy loss (F. Liu & Wang, 2022).

$$L_{BCE+DC}(y, \hat{p}) = L_{BCE}(y, \hat{p}) + L_{DC}(y, \hat{p}) \qquad (3)$$

Where $y$ is the ground truth and $\hat{p}$ is the model's estimated probability.

The Focal Tversky loss is a modified focal loss function built upon the Tversky index (Equation 4) designed to mitigate data imbalance. This loss function, expressed in Equation 5, is particularly effective at handling class imbalance by emphasizing challenging-to-segment regions within images. Research has demonstrated that employing the Focal Tversky loss can enhance segmentation accuracy, especially in scenarios where regions of interest are relatively small in comparison to the image size (Abraham & Khan, 2018).

$$TI = \frac{TP}{TP + \alpha FN + \beta FP} \tag{4}$$

Where TP is True positives, FN is False Negatives, and FP is False positives and incase $\alpha = \beta = 0.5$, it simplifies to dice coefficient.

$$FTL = (1 - TI)^{\gamma} \tag{5}$$

Where TI is the Tversky Index and $\gamma$ controls the non-linearity of the loss

**PaveSAM vs SAM**

The overlap between the ground truth and segmentation results of SAM and PaveSAM were evaluated using Dice Similarity Coefficient(DSC). The DSC is defined as :

$$DSC(A, B) = \frac{2(A \cap B)}{A + B}$$

where $\cap$ is the intersection(Zou et al., 2004).

Table 3 presents the DSC values for SAM and PaveSAM using our test data. To assess the effectiveness of PaveSAM across various pavement distresses, PaveSAM was evaluated using 15 randomly selected images. There was 31% enhancement in DSC across these 15 images in the test dataset. Furthermore, when PaveSAM was applied to all test images, there was a notable 34.9% improvement in the Dice score, detailed in Table 2.
Additionally, 10 images with distinct pavement distresses were randomly selected from the public dataset (crack500) and evaluated using PaveSAM, comparing the results with SAM. The DSC values for PaveSAM and SAM on crack500 are shown in Table 4.
When the PaveSAM and SAM models were tested on the entire test dataset, PaveSAM outperformed SAM by a significant margin. Specifically, PaveSAM demonstrated a 35% improvement over SAM when evaluated on the 10 images from crack500 and a remarkable 215% increase in performance across the entire dataset.

*Table 2: Dice score Comparison between SAM and PaveSAM on entire test data*

|  | PaveSAM Dice score | SAM Dice score |
|---|---|---|
| Our test data | 0.703 | 0.521 |
| Crack500 test data | 0.432 | 0.137 |

*Table 3: Dice Similarity Score Comparison between SAM and PaveSAM on 15 images our test data*

| *Pavement Distress* | *PaveSAM* | *SAM* | *Improvement* |
|---|---|---|---|
| *Longitudinal Crack* | 0.7773 | 0.5808 | 0.1965 |
| *Block crack* | 0.4645 | 0.1445 | 0.32 |
| *Longitudinal Crack* | 0.7179 | 0.1872 | 0.5307 |
| *Longitudinal Crack* | 0.7186 | 0.3083 | 0.4103 |
| *Block* | 0.5607 | 0.1195 | 0.4412 |
| *Transverse Crack* | 0.6809 | 0.227 | 0.4539 |
| *Transverse Crack* | 0.8148 | 0.7559 | 0.0589 |
| *Transverse Crack* | 0.7387 | 0.5553 | 0.1834 |
| *Alligator* | 0.6085 | 0.5793 | 0.0292 |
| *Block crack* | 0.6672 | 0.3708 | 0.2964 |
| *Block crack* | 0.3636 | 0.0866 | 0.277 |
| *Longitudinal Crack* | 0.7448 | 0.383 | 0.3618 |
| *Transverse Crack* | 0.6580 | 0.2561 | 0.4019 |
| *Longitudinal Crack* | 0.4062 | 0.0704 | 0.3358 |
| *Longitudinal Crack* | 0.5170 | 0.138 | 0.379 |
| | Average | | 0.311 |

*Table 4: Dice Similarity Score Comparison between SAM and PaveSAM on our crack500 data*

| Pavement Distress | PaveSAM | SAM | Improvement |
|---|---|---|---|
| *Longitudinal Crack* | 0.6156 | 0.0001 | 0.6155 |
| *Longitudinal Crack* | 0.7760 | 0.2291 | 0.5469 |
| *Transverse Crack* | 0.7140 | 0.4374 | 0.2766 |
| *Block* | 0.3172 | 0.1180 | 0.1992 |
| *Longitudinal Crack* | 0.6446 | 0.1468 | 0.4978 |
| *Longitudinal Crack* | 0.5124 | 0.1201 | 0.3923 |
| *Longitudinal Crack* | 0.7740 | 0.1584 | 0.6156 |
| *Longitudinal Crack* | 0.3224 | 0.1828 | 0.1396 |
| *Block* | 0.2164 | 0.1349 | 0.0815 |
| *Block* | 0.5196 | 0.2329 | 0.2867 |
| *Block* | 0.464 | 0.1870 | 0.277 |
| | Average | | 0.35715455 |

Figure 4 and Figure 5 compare the segmentation results of PaveSAM and SAM on our test data and crack500 dataset, respectively. In summary, fine-tuning SAM on pavement distresses improved its segmentation performance for pavement distresses, which are challenging for the pre-trained SAM due to varying shapes and width of pavement distresses.

| Image | Ground Truth | SAM | PaveSAM |
|---|---|---|---|

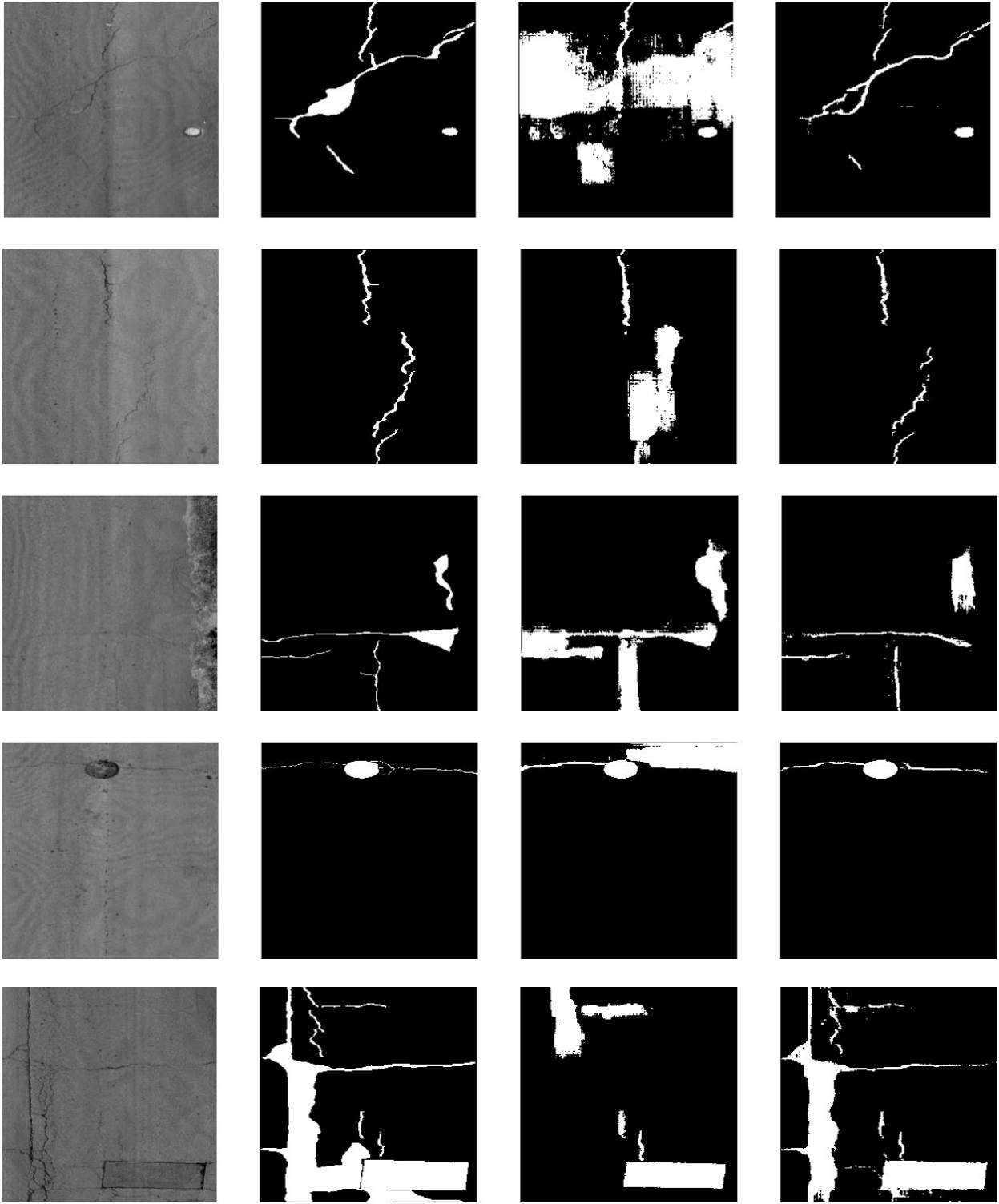

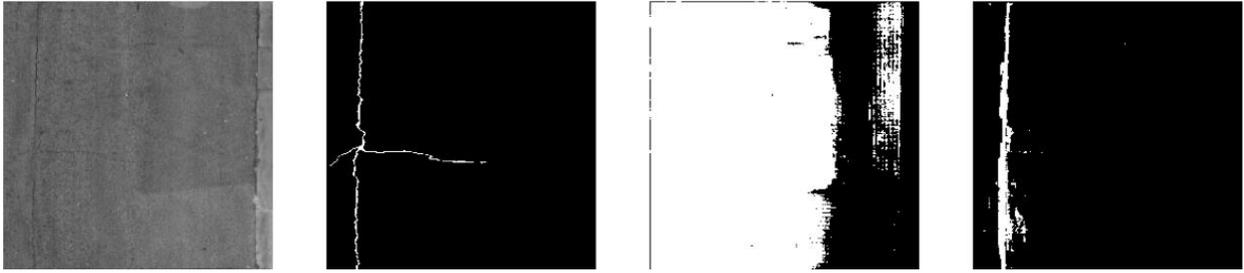

*Figure 4: Results of SAM model and PaveSAM on our test data*

| Image | Ground Truth | SAM | PaveSAM |
|-------|--------------|-----|---------|

*Figure 5: Segmentation results of SAM and PaveSAM on Crack500 dataset*

## Comparison with other models

The PaveSAM results underwent a rigorous comparative analysis against cutting-edge segmentation models, including UNet, UNet++, Transresunet, ResUNet++, FPN, Manet, Linknet, PSPNet, and DeepLab. UNet, distinguished as an encoder-decoder architecture, uniquely offers the flexibility to interchange its encoder with various Convolutional Neural Network (CNN) models. The primary objective of the UNet's encoder is to extract high-level features from the input image, aligning with the fundamental feature extraction goal shared by other prominent CNN architectures.

To enhance the versatility of our experimentation, a diverse set of CNN models were employed as the encoder for UNet. This encompassed a spectrum of influential architectures, such as ResNet50, ResNet152, MobileNetv2, VGG19, and EfficientNetB0. This strategic approach allowed us to explore and harness the distinctive feature extraction capabilities embedded in each CNN model, thereby providing a comprehensive evaluation of PaveSAM's performance across a wide spectrum of encoder choices within the UNet framework.

The models were trained with an input size of 256 x 256 x 3, utilizing a learning rate of 0.0001 and employing the Adam optimizer. The training spanned 100 epochs, conducted on the powerful Nvidia GeForce RTX 4080.

### U-Net
U-Net (Ronneberger et al., 2015) is a CNN based on a fully convolutional neural network where its architecture is altered and expanded to work with fewer training images to obtain significantly precise segmentation results. U-Net's architecture consists of a contracting path(encoder) to absorb context and a symmetric expansive path to facilitate precise localization. The contracting path follows the typical convolutional network with multiple convolutions accompanied by Rectified Linear Unit (ReLU) and max pooling operation. Similarly, the contracting part undergoes reduction in spatial information and increment in features information. However, the expansive path(decoder) integrates spatial and feature information using upconvolutions with feature information from the contracting path.

### VGG-19
VGG-19 is a convolutional neural network consisting of 47 layers. The network contains 19 layers with learnable weights consisting of 16 convolutional layers, and 3 fully connected layers (Simonyan & Zisserman, 2014). The pretrained version of VGG-19 trained on Imagenet is implemented  for this study.

### Resnet
Resnet-50 and Resnet-152 were the two backbones within the Resnet (He et al., 2015) family that were used in this study. Residual learning can come handy while adding shortcut connections to perform identity mapping, where the outputs are appended to the stacked layers output.

**UNet++**

UNet++ is a more robust architecture for segmentation, featuring nested, dense skip pathways to bridge semantic gaps between encoder and decoder feature maps. Comparative evaluations across various segmentation tasks demonstrate that UNet++ with deep supervision consistently outperforms U-Net and wide U-Net, achieving substantial improvements in Intersection over Union (IoU) scores(Zhou et al., 2018).

**ResUNet++**

ResUNet++ builds upon the Deep Residual U-Net (ResUNet) architecture by incorporating features like residual blocks, squeeze-and-excitation blocks, ASPP (Atrous Spatial Pyramid Pooling), and attention blocks. This architecture consists of a stem block, three encoder blocks, ASPP module, and three decoder blocks, leveraging these components to enhance feature extraction and segmentation (Jha et al., 2019).

Transresunet

Transformer ResUNet (TransResU-Net)(Tomar et al., 2022) for automated segmentation utilizes residual blocks with ResNet-50 as the base and integrates transformer self-attention mechanisms along with dilated convolutions to enhance performance. Experimental results on benchmark datasets demonstrate TransResU-Net's promising dice score and real-time processing speed.

**EfficientNet**

EfficientNet (Tan & Le, 2019) is a CNN and scaling based network that uniformly scales overall resolution, width and depth dimensions through compound coefficients. The compound scaling technique is justified over the premise that whenever the size of an input image is larger, the network requires additional layers to increase the receptive field and extra channels to absorb more fine-grained patterns from the larger image. The EfficientNet-B0 backbone used in this study is based on MobileNetV2's inverted bottleneck residual blocks in conjunction with the squeeze and excitation blocks.

**Feature Pyramid Networks(FPN)**

Feature Pyramid Networks (FPN) for object detection is a neural network architecture designed to address the challenge of detecting objects at various scales within an image (Lin et al., 2016). FPN incorporates a top-down pathway with lateral connections, creating a feature pyramid that combines high-level semantic information from deeper layers with fine-grained spatial details from shallower layers. This design ensures that the network captures multi-scale features, making it effective for detecting objects of different sizes.

**DeepLab**

DeepLab is a deep learning-based semantic segmentation model designed for high-precision image understanding(Chen et al., 2016). Introduced by Google, DeepLab employs a deep convolutional neural network (CNN) architecture to segment images into various object classes and assign pixel-wise labels. One of its notable features is the use of atrous convolution (also known as dilated convolution), which allows the network to capture multi-scale contextual information without downsampling the input resolution. In this study, Deeplabv2 was implemented.

**Pyramid Scene Parsing Network**

PSPNet, or Pyramid Scene Parsing Network (Zhao et al., 2016), is a convolutional neural network architecture designed for semantic image segmentation. Introduced to address the challenge of capturing global contextual information, PSPNet employs a pyramid pooling module to aggregate features at different scales. This module allows the network to consider both local and global contexts, enhancing its ability to accurately classify pixels in an image. The pyramid structure facilitates the integration of information from varying receptive fields, enabling PSPNet to capture intricate details and context-aware representations.

**MobileNet**

MobileNet is a family of lightweight deep learning models designed for efficient deployment on mobile and edge devices with limited computational resources(Howard et al., 2017). Introduced by Google, MobileNet architectures leverage depthwise separable convolutions to reduce the number of parameters and computations, making them well-suited for real-time applications on devices with constraints such as smartphones and IoT devices. The use of separable convolutions enables a balance between model accuracy and computational efficiency, making MobileNet particularly popular for tasks like image classification, object detection, and semantic segmentation on resource-constrained platforms. The architecture's adaptability and efficiency have contributed to its widespread adoption in various mobile and embedded applications.

**LinkNet**

LinkNet (Chaurasia & Culurciello, 2017) is a deep learning architecture designed for efficient semantic segmentation by leveraging encoder representations. The model focuses on optimizing the segmentation process by incorporating encoder features effectively. Through its design, LinkNet aims to balance accuracy and computational efficiency in semantic segmentation tasks, where the goal is to classify each pixel in an image into specific categories. The utilization of encoder representations enhances the network's ability to capture relevant features and contextual information, contributing to improved segmentation results. This approach is particularly valuable for real-time applications and scenarios where computational resources are constrained, making LinkNet a notable contribution to the field of semantic segmentation in computer vision.

**MAnet**

MAnet stands for mask attention mechanism (MA) which uses the predicted masks at the first stage as spatial attention maps to adjust the features of the CNN at the second stage. The MA spatial attention maps for features calculate the percentage of masked pixels in their receptive fields, suppressing the feature values based on the overlapping rates between their receptive fields and the segmented regions(Xu et al., 2021).

**Evaluation metrics**

In this study, four widely recognized evaluation metrics were employed to objectively gauge the efficacy of the proposed approach and accurately assess the performance of the network model. Our quantitative evaluation encompassed the F1 score, Recall, Precision, and Intersection over Union (IoU).

Precision (P), defined by Eq. (4), serves as the ratio between relevant instances and retrieved instances. It measures the accuracy of classification by determining the proportion of correctly identified observations per predicted class. On the other hand, Recall (R), as per Eq. (5), signifies the ratio of relevant instances successfully retrieved. It is a critical measure of sensitivity, representing the percentage of actual positives correctly identified. It is important to note the inherent trade-off between Precision and Recall; an increase in precision may lead to a decrease in model sensitivity, and vice versa. Striking a balance between these two indicators is crucial to achieving a model that optimally fits the input data(Di Benedetto et al., 2023).

$$Precision = \frac{TP}{TP+FP} \qquad (4)$$

$$Recall = \frac{TP}{TP+FN} \qquad (5)$$

Where TP is True Positives, FP is False Positives and FN is False Negatives

The F1 score (Eq. 6) is a harmonic mean of Precision and Recall, provides a consolidated assessment of model performance. It proves to be a robust indicator suitable for both balanced and unbalanced datasets. F1 scores exceeding 0.9 suggest highly accurate classification, while scores below 0.5 indicate potential inaccuracies, rendering the classification unsuitable. Analyzing F1 becomes imperative when seeking a harmonious equilibrium between Precision and Recall.

$$F1\ score\ =\ 2\ x\ \frac{Precision*Recall}{Precision+Recall} \qquad (6)$$

The Intersection over Union (IoU), a geometric evaluation metric, quantifies the proximity of predicted results to the ground-truth. Mean Intersection over Union (MIoU), computed by Eq. (6), establishes the ratio between the intersection and union of two sets: the ground truth and the predicted segmentation. The predicted results represent the mask generated by the proposed model, while the ground-truth corresponds to the mask used for training. IoU is thereby calculated as the common pixels between the target and predicted masks divided by the total pixels in both masks, encapsulating the overlap between the two. This geometric evaluation provides insights into the accuracy of segmentation and the model's fidelity in reproducing the ground truth (F. Liu & Wang, 2022).

$$IOU = \frac{1}{k+1}\sum_{i=0}^{k}\frac{p_{ii}}{\sum_{j=0}^{k}p_{ij}+\sum_{j=0}^{k}p_{ji}-p_{ii}} \qquad (7)$$

Where $p_{ii}$ is the number of true positives , $p_{ij}$ is the number of false positives and $p_{ji}$ is the number of false negatives, k+1 classes is the number of classes including the background.

## RESULTS AND DISCUSSION

## Model accuracy

Table 9 presents the accuracy results of the models, utilizing a uniform threshold of 0.5 across all instances. PaveSAM emerged as the best performer across multiple metrics, including Precision, Recall, F1 score, and IOU Score. Following closely to PaveSAM, UNet-VGG19 secured the second position in all assessment criteria except IOU Score, where it was surpassed. Specifically, in Precision, UNet-VGG19 was followed by UNet++, LinkNet and UNet-ResNet50.

In terms of Recall, UNet-VGG19 was, succeeded by Transresunet, UNet++, UNet-ResNet50, UNetEfficientb0, and LinkNet. F1 Score rankings placed UNet-VGG19 at the second, trailed by Transresunet, Unet++, UNet-ResNet50, LinkNet, and ResNet-152. Shifting focus to IOU Score, UNet-Efficientb0 claimed the second position, trailing UNet-VGG19, and outperforming Transresunet, UNet++, UNet-ResNet50 and Linknet.

These comparative evaluations are graphically depicted in Figure 6, providing a visual representation of the models' performance across diverse metrics. The generated masks for the top four models on the test data are visually presented in Figure 9. The Dice scores and IOU of the top models during training is shown in Figure 7 and Figure 8.

*Table 5: Evaluation for the Models*

|  |  | *Precision* | *Recall* | *F1 score* | *IOU score* |
|---|---|---|---|---|---|
| *PaveSAM* |  | 0.714 | 0.764 | 0.703 | 0.578 |
| *UNet++-ResNet50* | (Zhou et al., 2018) | 0.695 | 0.497 | 0.542 | 0.44 |
| *Res-Unet++* | (Jha et al., 2019) | 0.613 | 0.4401 | 0.442 | 0.343 |
| *Transresunet* | (Tomar et al., 2022) | 0.677 | 0.536 | 0.566 | 0.457 |
| *UNet-ResNet50* | (Ronneberger et al., 2015) | 0.684 | 0.495 | 0.522 | 0.426 |
| *UNet-ResNet152* | (Ronneberger et al., 2015) | 0.658 | 0.480 | 0.514 | 0.416 |
| *UNet-MobileNetv2* | (Howard et al., 2017). | 0.602 | 0.479 | 0.499 | 0.401 |
| *UNet-Efficientb0* | (Tan & Le, 2019) | 0.600 | 0.494 | 0.400 | 0.496 |
| *UNet-VGG19* | (Simonyan & Zisserman, 2014). | 0.700 | 0.563 | 0.585 | 0.472 |
| *FPN-ResNet50* | (Lin et al., 2016). | 0.601 | 0.4255 | 0.447 | 0.364 |
| ***DeepLabv3-ResNet50*** | Chen et al., 2016) | 0.609 | 0.438 | 0.465 | 0.382 |
| ***Manet-ResNet50*** | (Xu et al., 2021). | 0.671 | 0.401 | 0.430 | 0.347 |
| ***PSPNet*** | (Zhao et al., 2016) | 0.591 | 0.427 | 0.446 | 0.360 |
| ***LinkNet*** | (Chaurasia & Culurciello, 2017) | 0.691 | 0.482 | 0.517 | 0.418 |

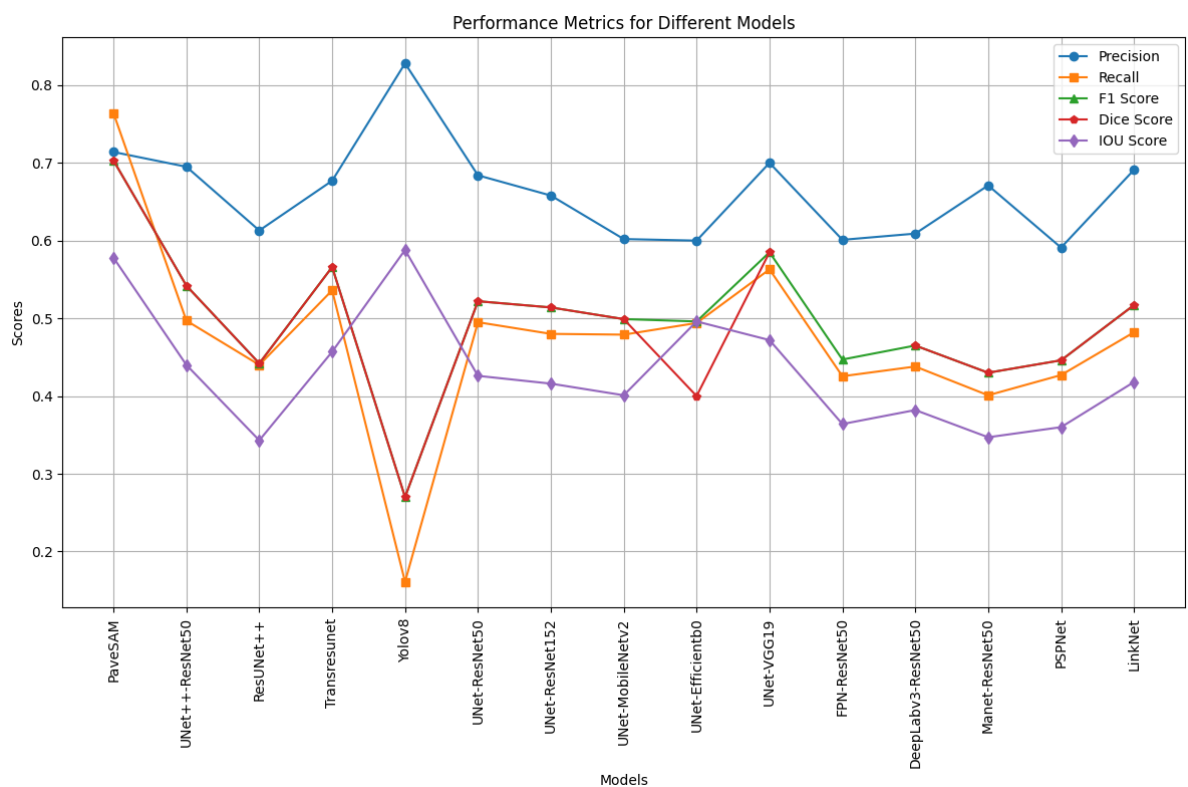

*Figure 6: Comparison of the evaluation metrics*

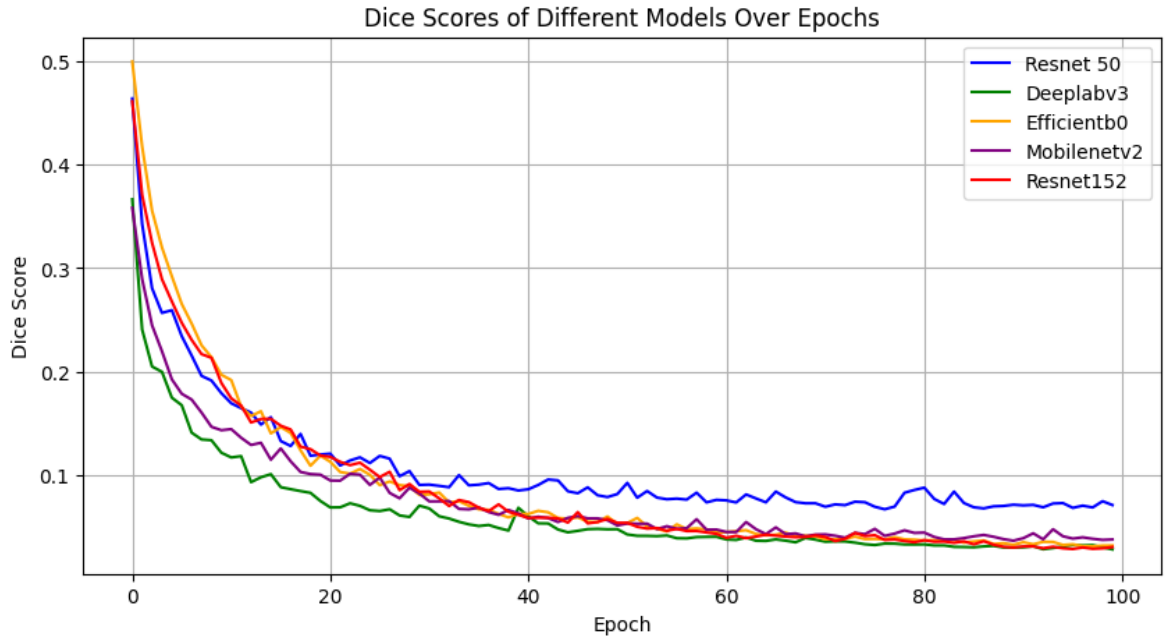

*Figure 7: The Dice scores of the models during training*

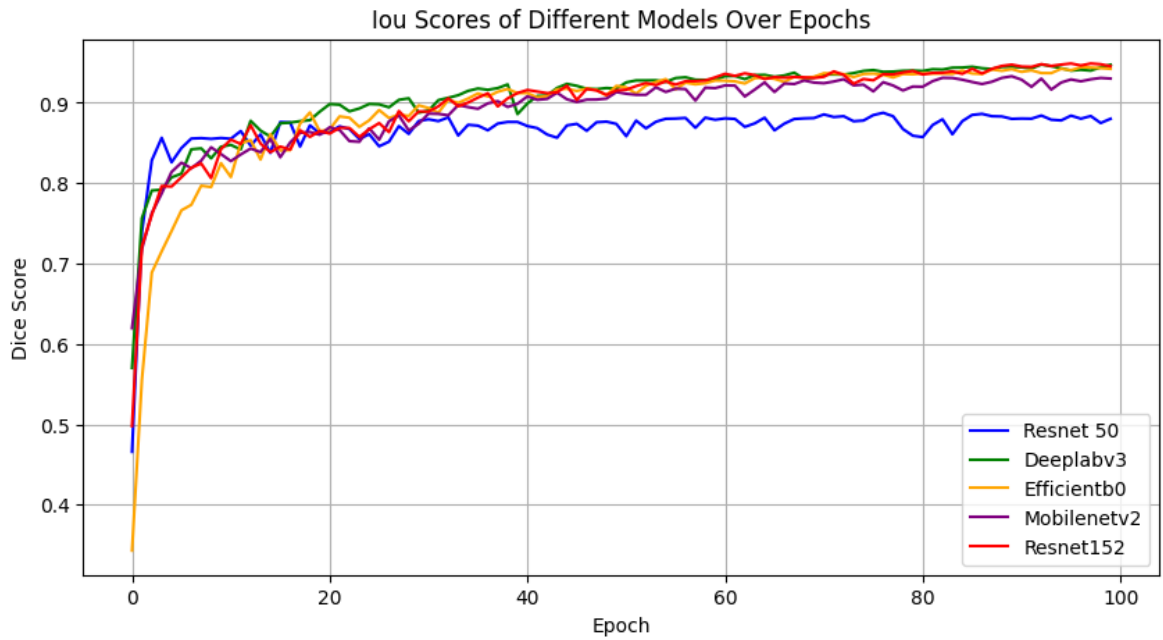

*Figure 8: IOU Scores of the models during training*

| Original Image | Ground Mask | Truth | UNet++-Resnet50 | UNet-Vgg19 | Transresunet | PaveSAM |
|---|---|---|---|---|---|---|

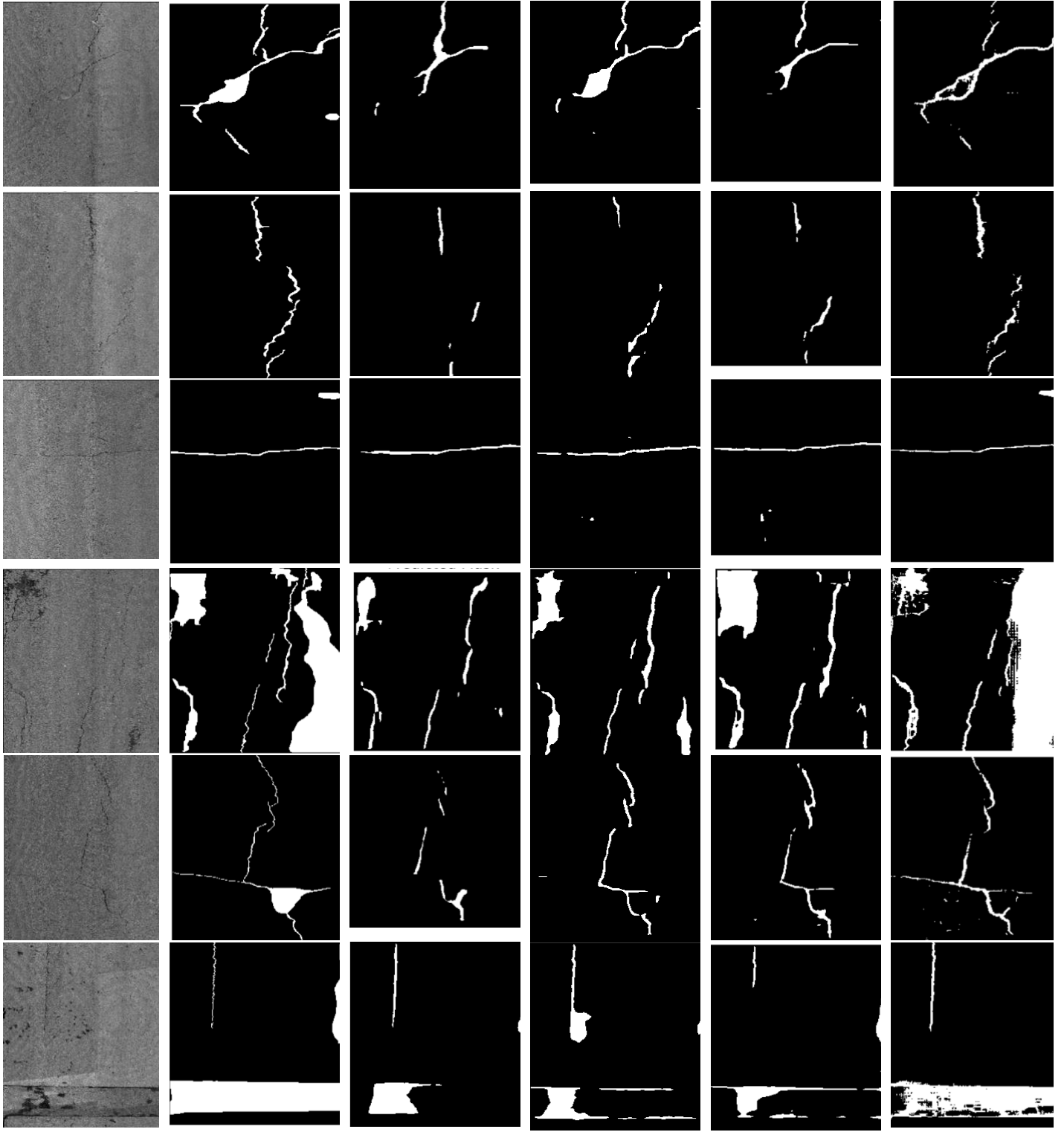

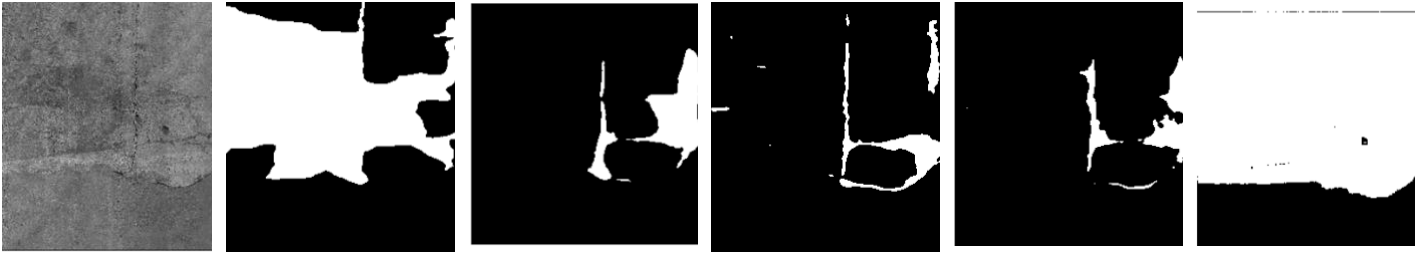

*Figure 9: Segmentation for the test images*

**Crack500 dataset results:**

Initially, the PaveSAM model was trained on our dataset and subsequently evaluated its performance on our dataset and on Crack500 dataset. After the PaveSAM model trained on Crack500 was evaluated in comparison with other state-of-the-art models also trained and tested on the same dataset, presenting the results in Table 6 and Figure 10. The PaveSAM model was trained using a combination of Dice and Binary Cross Entropy loss, as well as Focal Tversky loss, and notably outperformed all other state-of-the-art models in terms of recall, F1 score, and IOU score. UNet-VGG19 demonstrated the highest precision among all models, followed by PaveSAM, UNet++, and UNet-MobileNetv2. The SAM model exhibited the lowest scores in F1 score, precision, and IOU score, highlighting the substantial performance improvements achieved by PaveSAM through fine-tuning for pavement distress segmentation. The generated masks for the top four models on the test data are visually presented in Figure 11.

*Table 6: Evaluation of Models trained and teste on crack500*

| | | Precision | Recall | F1 score | IOU score |
|---|---|---|---|---|---|
| **PaveSAM (Dice +BCE loss)** | | 0.696 | 0.723 | 0.691 | 0.538 |
| **PaveSAM (Focal Tversky loss)** | | 0.730 | 0.668 | 0.676 | 0.522 |
| **SAM** | (Kirillov et al., 2023) | 0.105 | 0.672 | 0.148 | 0.095 |
| **UNet++-ResNet50** | (Zhou et al., 2018) | 0.725 | 0.624 | 0.651 | 0.496 |
| **Res-Unet++** | (Jha et al., 2019) | 0.689 | 0.5722 | 0.583 | 0.436 |
| **Transresunet** | (Tomar et al., 2022) | 0.713 | 0.645 | 0.659 | 0.504 |
| ***UNet-ResNet50*** | (Ronneberger et al., 2015) | 0.704 | 0.609 | 0.624 | 0.469 |
| ***UNet-ResNet152*** | (Ronneberger et al., 2015) | 0.702 | 0.588 | 0.617 | 0.467 |
| ***UNet-MobileNetv2*** | (Howard et al., 2017). | 0.720 | 0.09 | 0.16 | 0.09 |
| ***UNet-Efficientb0*** | (Tan & Le, 2019) | 0.706 | 0.596 | 0.614 | 0.463 |
| ***UNet-VGG19*** | (Simonyan & Zisserman, 2014). | 0.75 | 0.572 | 0.62 | 0.470 |
| ***DeepLabv3+-ResNet50*** | (Chen et al., 2018) | 0.655 | 0.488 | 0.504 | 0.370 |
| ***Manet-ResNet50*** | (Xu et al., 2021). | 0.703 | 0.597 | 0.615 | 0.466 |
| ***LinkNet*** | (Chaurasia & Culurciello, 2017) | 0.681 | 0.602 | 0.613 | 0.459 |

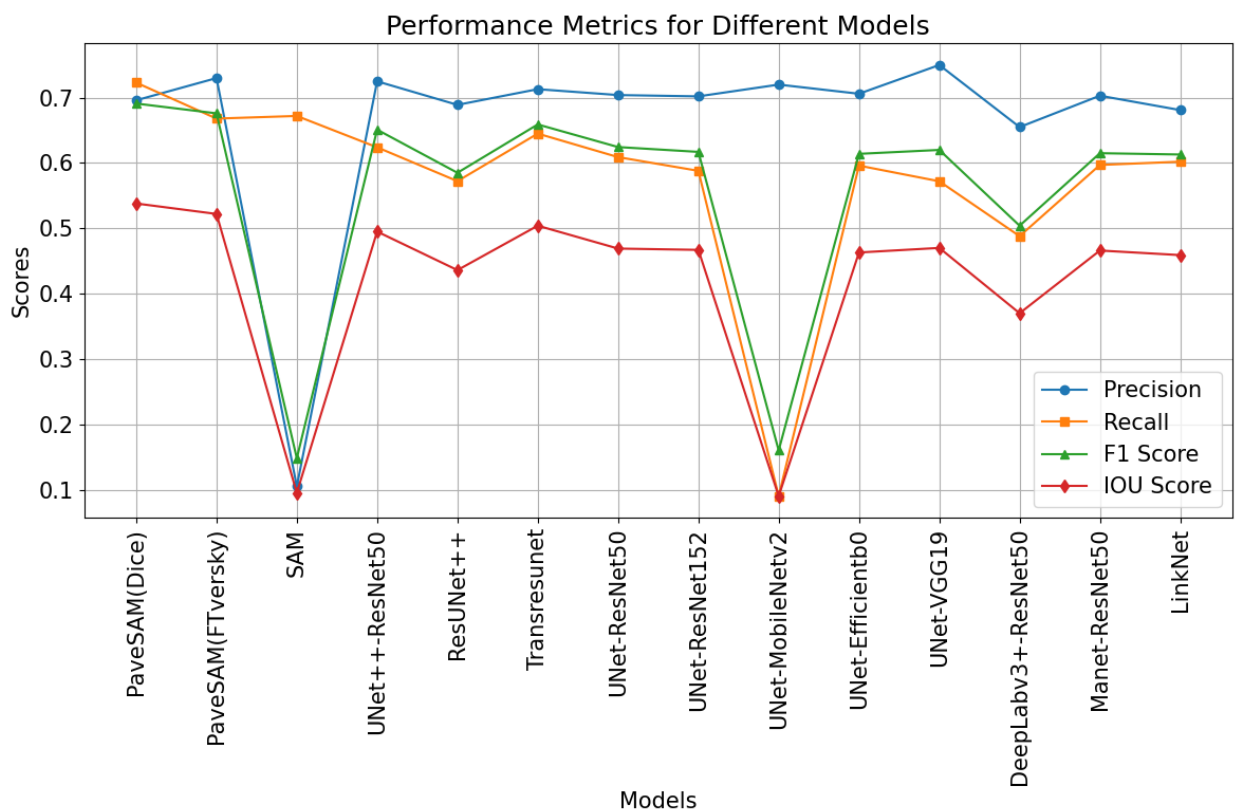

*Figure 10: Performance Metrics of the various models tested on crack500*

| Original Image | Ground Truth Mask | Transresunet | UNet-Resnet50 | UNet++ | PaveSAM |
|---|---|---|---|---|---|

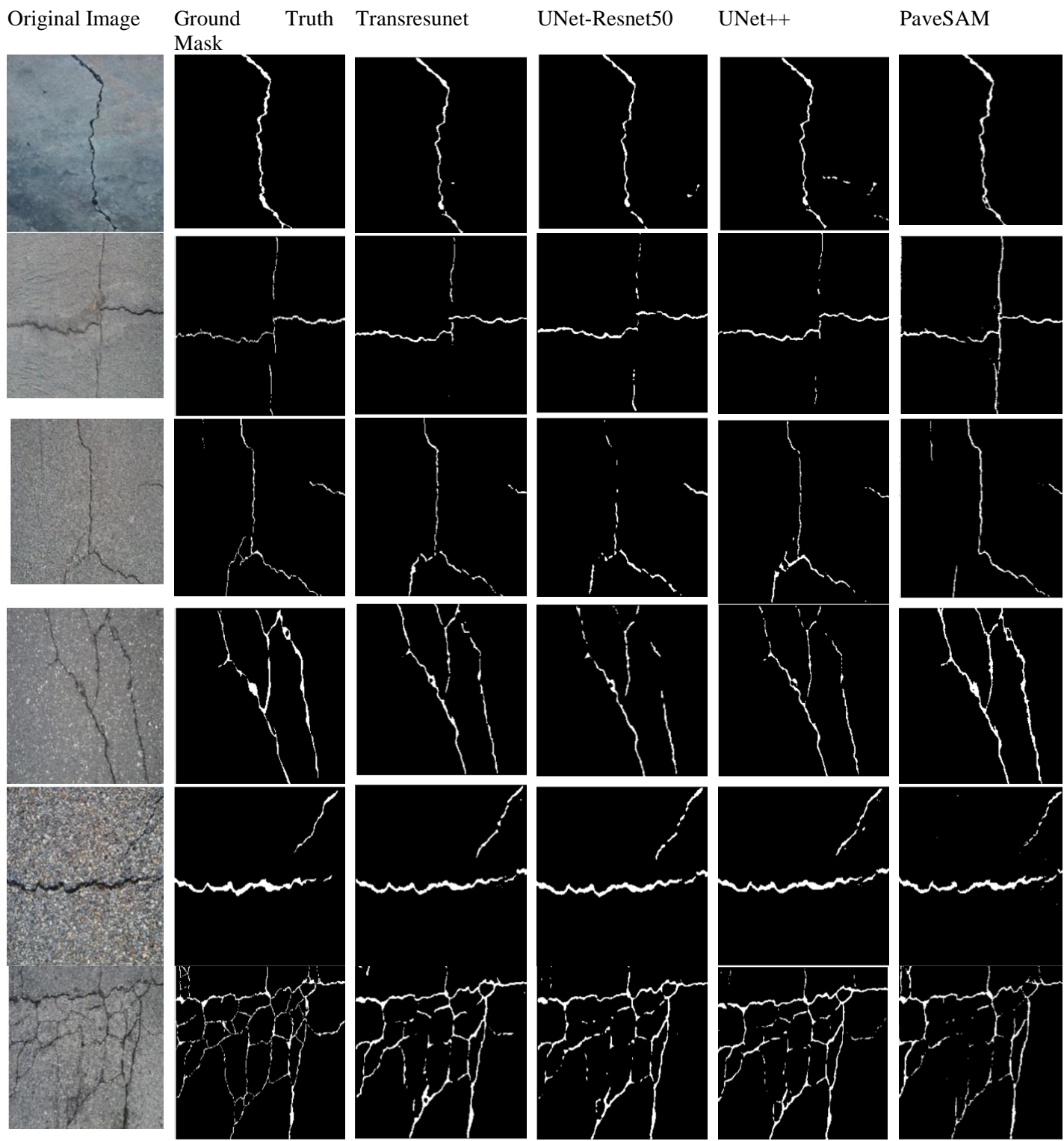

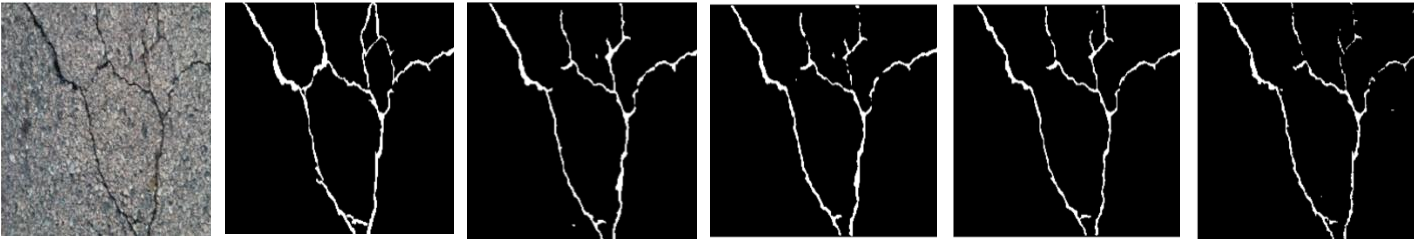

*Figure 11: Segmentation on crack500 dataset*

**Model Complexity and Speed**
**Learnable parameters**

Analyzing model complexity(see Table 7 and Figure 12), the number of learnable parameters were investigated. The ResUnet++ model has the fewest parameters (4.06M), followed by UNet-Efficientb0 (6.25M), and then UNet-Mobilenetv2. Conversely, the PaveSAM model is the most parameter-heavy (136M), with MA-Net (147.4M) and DeepLabv3 (39.63M) also showing high parameter counts. For PaveSAM, the encoder accounts for 132 million parameters, the mask decoder 3.87 million, and the prompt encoder 0.006 million. Only the PaveSAM mask decoder was trained, comprising 3.87M parameters.

**Model complexity**
Furthermore, computational complexity(as shown in Table 7) was measured in Floating-Point Operations (FLOPs). PaveSAM exhibited the highest FLOPS at 487 GFLOPs, followed by UNet-Efficientb0 (125M) and ResUNet++ with the lowest GFLOPS at 15.85, followed by PSPNet (18.62 GFLOPS).

**Model Speed**
Regarding frames per second (FPS), Unet-VGG-19(183 FPS) was the fastest, followed by ResUnet++ (175 FPS) and UNet-Resnet50(173 FPS). PaveSAM had the lowest FPS at 6.28 as shown in Table 5. Breaking down PaveSAM's FPS performance, the mask encoder operated at 511 FPS, the prompt encoder at 2336 FPS, and the image encoder at 6.4 FPS. The image encoder significantly contributes to the high GFLOPS, parameter count, and low FPS of the PaveSAM model. The model speed were tested on Nvidia GeForce RTX 4080.

The objective of this paper was to fine-tune SAM for pavement distress segmentation, as it struggled with segmenting tiny and irregular pavement distresses, unlike the natural images it was trained on. PaveSAM excelled in segmenting pavement distress on our dataset and Crack500, surpassing current state-of-the-art models. However, PaveSAM's drawback lies in its large and complex nature. Future work aims to develop a Tiny PaveSAM that is computationally efficient and suitable for edge devices.

*Table 7: Memory complexity of different models*

|  | GFLOPs | Parameters(M) | Model size(MB) |
|---|---|---|---|
| **UNet-ResNet50** | 24.93 | 32.5 | 130.5 |

| Model | | | |
|---|---|---|---|
| **PaveSAM** | 487 | 136M (3.87M trained) | 375.1 |
| **UNet++-Res50** | 75.15 | 48.9 | 196.4 |
| **ResUnet++** | 15.85 | 4.06 | 16.4 |
| **Transresunet** | 24.13 | 27.6 | 110.8 |
| **Yolov8** | 42.6 | 11.8 | 23 |
| **UNet-ResNet152** | 51.46 | 67.15 | 269.7 |
| **UNet-MobileNetv2** | 36.20 | 6.62 | 26.8 |
| **UNet-Efficientb0** | 125.97 | 6.25 | 25.4 |
| **UNet-VGG19** | 44.6 | 29.05 | 116.3 |
| **FPN-ResNet50** | 40.03 | 26.11 | 104.8 |
| **DeepLabv3-ResNet50** | 30.40 | 39.63 | 158.9 |
| **Manet-ResNet50** | 113.18 | 147.44 | 590.2 |
| **PSPNet** | 18.62 | 24.3 | 97.6 |
| **Linknet** | 19.62 | 31.17 | 125.1 |

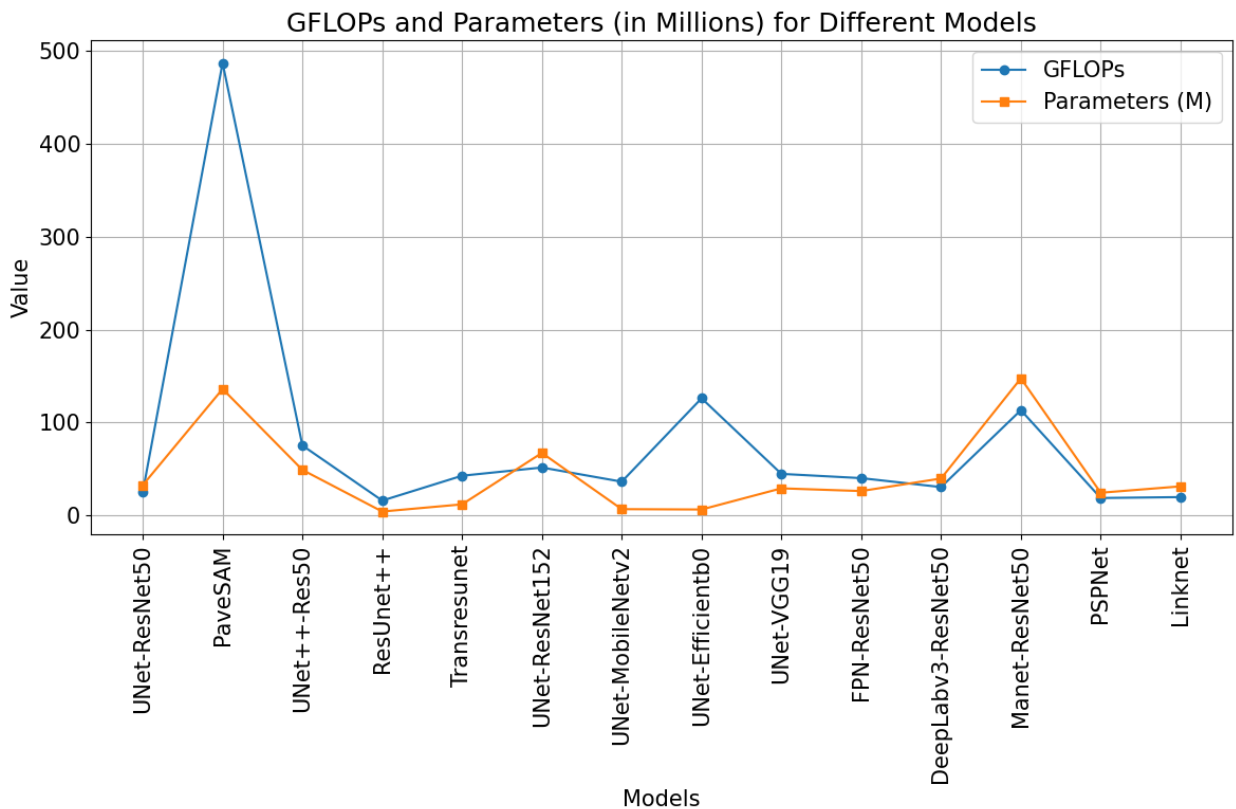

Figure 12: Model complexity of different models

Table 8: FPS for the different models

| | **FPS** |
|---|---|
| *PaveSAM* | 6.28 |
| *UNet++-ResNet50* | 81 |
| *Res-Unet++* | 175 |
| *Transresunet* | 120 |
| *UNet-ResNet50* | 173 |
| *UNet-ResNet152* | 87.64 |
| *UNet-MobileNetv2* | 154 |
| *UNet-Efficientb0* | 117 |
| *UNet-VGG19* | 183 |
| *DeepLabv3+-ResNet50* | 166 |
| *Manet-ResNet50* | 101 |
| *LinkNet* | 162 |

**Ethical considerations:**

The datasets used in this research are publicly available and licensed for academic use, thereby eliminating any data privacy concerns associated with their usage.

**CONCLUSION**

In summary, our research tackles the challenges of automated pavement distress analysis by introducing PaveSAM, a well-adjusted version of the zero-shot segmentation model SAM. PaveSAM performs exceptionally well, as confirmed by tests on different datasets, showing its potential to improve how we analyze pavement issues.

Our unique method of using bounding box prompts for zero-shot segmentation turns out to be a smart way to reduce the effort and cost of labeling compared to traditional methods. PaveSAM also stands out for its ability to work well with only a small set of 180 images, making it practical for situations with limited data.

Furthermore, our contribution extends to the wider community involved in pavement distress analysis. PaveSAM can effectively use existing datasets that have bounding box annotations, making it a versatile tool for various pavement distress databases. This innovation opens up new possibilities for more accessible and diverse segmentation datasets.

In essence, our research not only improves an advanced model for pavement distress analysis but also introduces a method that changes how automated pavement monitoring is approached, promising a more efficient and effective way to analyze distress.

To calculate the Pavement Condition Index (PCI), one needs detailed information on pavement distress including type, severity, and extent, which can be automatically calculated through pixel-level segmentation. Most segmentation models are limited to images they were trained on, requiring a large number of annotated images for accurate results, but PaveSAM offers zero-shot segmentation and can be integrated into automated pavement management systems. This means that our model can be applied to any existing pavement distress database without requiring training which no current segmentation model can do.

In future studies, an alternative approach could involve utilizing text prompts for pavement distress segmentation instead of relying on bounding box prompts. By implementing text prompts, the model could be adapted for use with datasets lacking pre-existing bounding boxes. Additionally, future work may focus on reducing the model size to enhance its suitability for deployment on edge devices. These advancements could significantly broaden the applicability and efficiency of the model in practical scenarios.

**Limitations:**

The model relies on bounding box annotations to segment images, which means it can only be applied to datasets with pre-existing bounding boxes or requires passing images through a detection model to generate them. Moreover, the distribution of pavement distresses in the training data was uneven, potentially biasing the model towards more dominant distress types.
Additionally, the model was trained using top-down images from Laser Crack Measurement Systems (LCMS), suggesting potential limitations when applied to images from other orientations, such as those captured from dash cameras.

# REFERENCES

Abraham, N., & Khan, N. M. (2018). *A Novel Focal Tversky loss function with improved Attention U-Net for lesion segmentation*. https://doi.org/10.48550/ARXIV.1810.07842

Bucher, M., Vu, T.-H., Cord, M., & Pérez, P. (2019). *Zero-Shot Semantic Segmentation* (arXiv:1906.00817). arXiv. http://arxiv.org/abs/1906.00817

Chaurasia, A., & Culurciello, E. (2017). *LinkNet: Exploiting Encoder Representations for Efficient Semantic Segmentation*. https://doi.org/10.48550/ARXIV.1707.03718

Chen, L.-C., Papandreou, G., Kokkinos, I., Murphy, K., & Yuille, A. L. (2016). *DeepLab: Semantic Image Segmentation with Deep Convolutional Nets, Atrous Convolution, and Fully Connected CRFs*. https://doi.org/10.48550/ARXIV.1606.00915

Chen, L.-C., Zhu, Y., Papandreou, G., Schroff, F., & Adam, H. (2018). *Encoder-Decoder with Atrous Separable Convolution for Semantic Image Segmentation* (arXiv:1802.02611). arXiv. http://arxiv.org/abs/1802.02611

Chua, K. M., & Xu, L. (1994). Simple Procedure for Identifying Pavement Distresses from Video Images. *Journal of Transportation Engineering*, *120*(3), 412–431. https://doi.org/10.1061/(ASCE)0733-947X(1994)120:3(412)

Di Benedetto, A., Fiani, M., & Gujski, L. M. (2023). U-Net-Based CNN Architecture for Road Crack Segmentation. *Infrastructures*, *8*(5), 90. https://doi.org/10.3390/infrastructures8050090

Fu, K. S., & Mui, J. K. (1981). A survey on image segmentation. *Pattern Recognition*, *13*(1), 3–16. https://doi.org/10.1016/0031-3203(81)90028-5

Ghosh, R., & Smadi, O. (2021). Automated Detection and Classification of Pavement Distresses using 3D Pavement Surface Images and Deep Learning. *Transportation Research Record: Journal of the Transportation Research Board*, *2675*(9), 1359–1374. https://doi.org/10.1177/03611981211007481

Gonzalez, R. C., Woods, R. E., & Eddins, S. L. (2009). *Digital image processing using MATLAB* (5. impr). Pearson Education [u.a.].

Groenewald, A. M., Barnard, E., & Botha, E. C. (1993). Related approaches to gradient-based thresholding. *Pattern Recognition Letters*, *14*(7), 567–572. https://doi.org/10.1016/0167-8655(93)90107-O

Guan, S., Liu, H., Pourreza, H. R., & Mahyar, H. (2023). *Deep Learning Approaches in Pavement Distress Identification: A Review* (arXiv:2308.00828). arXiv. http://arxiv.org/abs/2308.00828

Han, C., Ma, T., Huyan, J., Huang, X., & Zhang, Y. (2022). CrackW-Net: A Novel Pavement Crack Image Segmentation Convolutional Neural Network. *IEEE Transactions on Intelligent Transportation Systems*, *23*(11), 22135–22144. https://doi.org/10.1109/TITS.2021.3095507

He, K., Chen, X., Xie, S., Li, Y., Dollár, P., & Girshick, R. (2021). *Masked Autoencoders Are Scalable Vision Learners*. https://doi.org/10.48550/ARXIV.2111.06377

He, K., Zhang, X., Ren, S., & Sun, J. (2015). *Deep Residual Learning for Image Recognition*. https://doi.org/10.48550/ARXIV.1512.03385

Howard, A. G., Zhu, M., Chen, B., Kalenichenko, D., Wang, W., Weyand, T., Andreetto, M., & Adam, H. (2017). *MobileNets: Efficient Convolutional Neural Networks for Mobile Vision Applications*. https://doi.org/10.48550/ARXIV.1704.04861

Iandola, F. N., Han, S., Moskewicz, M. W., Ashraf, K., Dally, W. J., & Keutzer, K. (2016). *SqueezeNet: AlexNet-level accuracy with 50x fewer parameters and <0.5MB model size*. https://doi.org/10.48550/ARXIV.1602.07360

Jha, D., Smedsrud, P. H., Riegler, M. A., Johansen, D., de Lange, T., Halvorsen, P., & Johansen, H. D. (2019). *ResUNet++: An Advanced Architecture for Medical Image Segmentation* (arXiv:1911.07067). arXiv. http://arxiv.org/abs/1911.07067

Jun, F., Jiakuan, L., Yichen, S., Ying, Z., & Chenyang, Z. (2022). ACAU-Net: Atrous Convolution and Attention U-Net Model for Pavement Crack Segmentation. *2022 International Conference on Computer Engineering and Artificial Intelligence (ICCEAI)*, 561–565. https://doi.org/10.1109/ICCEAI55464.2022.00120

Kang, J., & Feng, S. (2022). Pavement Cracks Segmentation Algorithm Based on Conditional Generative Adversarial Network. *Sensors*, *22*(21), Article 21. https://doi.org/10.3390/s22218478

Kheradmandi, N., & Mehranfar, V. (2022). A critical review and comparative study on image segmentation-based techniques for pavement crack detection. *Construction and Building Materials*, *321*, 126162. https://doi.org/10.1016/j.conbuildmat.2021.126162

Kirillov, A., Mintun, E., Ravi, N., Mao, H., Rolland, C., Gustafson, L., Xiao, T., Whitehead, S., Berg, A. C., Lo, W.-Y., Dollár, P., & Girshick, R. (2023). *Segment Anything* (arXiv:2304.02643). arXiv. http://arxiv.org/abs/2304.02643

Lau, S. L. H., Chong, E. K. P., Yang, X., & Wang, X. (2020). Automated Pavement Crack Segmentation Using U-Net-Based Convolutional Neural Network. *IEEE Access*, *8*, 114892–114899. https://doi.org/10.1109/ACCESS.2020.3003638

Li, D., Duan, Z., Hu, X., & Zhang, D. (2021). Pixel-Level Recognition of Pavement Distresses Based on U-Net. *Advances in Materials Science and Engineering*, *2021*, 1–11. https://doi.org/10.1155/2021/5586615

Lin, T.-Y., Dollár, P., Girshick, R., He, K., Hariharan, B., & Belongie, S. (2016). *Feature Pyramid Networks for Object Detection*. https://doi.org/10.48550/ARXIV.1612.03144

Liu, F., & Wang, L. (2022). UNet-based model for crack detection integrating visual explanations. *Construction and Building Materials*, *322*, 126265. https://doi.org/10.1016/j.conbuildmat.2021.126265

Liu, X., Wu, K., Cai, X., & Huang, W. (2023). Semi-supervised semantic segmentation using cross-consistency training for pavement crack detection. *Road Materials and Pavement Design*, 1–13. https://doi.org/10.1080/14680629.2023.2266853

Ma, J., He, Y., Li, F., Han, L., You, C., & Wang, B. (2023). *Segment Anything in Medical Images* (arXiv:2304.12306). arXiv. http://arxiv.org/abs/2304.12306

Majidifard, H., Adu-Gyamfi, Y., & Buttlar, W. G. (2020). Deep machine learning approach to develop a new asphalt pavement condition index. *Construction and Building Materials*, *247*, 118513. https://doi.org/10.1016/j.conbuildmat.2020.118513

Nguyen, T. S., Begot, S., Duculty, F., & Avila, M. (2011). Free-form anisotropy: A new method for crack detection on pavement surface images. *2011 18th IEEE International Conference on Image Processing*, 1069–1072. https://doi.org/10.1109/ICIP.2011.6115610

Nnolim, U. A. (2020). Automated crack segmentation via saturation channel thresholding, area classification and fusion of modified level set segmentation with Canny edge detection. *Heliyon*, *6*(12), e05748. https://doi.org/10.1016/j.heliyon.2020.e05748

Otsu, N. (1979). A Threshold Selection Method from Gray-Level Histograms. *IEEE Transactions on Systems, Man, and Cybernetics*, *9*(1), 62–66. https://doi.org/10.1109/TSMC.1979.4310076

Ouyang, A., & Wang, Y. (2012). *Edge Detection In Pavement Crack Image With Beamlet Transform*. 2036–2039. https://doi.org/10.2991/emeit.2012.451

Owor, N. J., Du, H., Daud, A., Aboah, A., & Adu-Gyamfi, Y. (2023). *Image2PCI -- A Multitask Learning Framework for Estimating Pavement Condition Indices Directly from Images* (arXiv:2310.08538). arXiv. http://arxiv.org/abs/2310.08538

Pourpanah, F., Abdar, M., Luo, Y., Zhou, X., Wang, R., Lim, C. P., Wang, X.-Z., & Wu, Q. M. J. (2022). A Review of Generalized Zero-Shot Learning Methods. *IEEE Transactions on Pattern Analysis and Machine Intelligence*, 1–20. https://doi.org/10.1109/TPAMI.2022.3191696

Redmon, J., Divvala, S., Girshick, R., & Farhadi, A. (2015). *You Only Look Once: Unified, Real-Time Object Detection*. https://doi.org/10.48550/ARXIV.1506.02640

Ronneberger, O., Fischer, P., & Brox, T. (2015). *U-Net: Convolutional Networks for Biomedical Image Segmentation*. https://doi.org/10.48550/ARXIV.1505.04597

Simonyan, K., & Zisserman, A. (2014). *Very Deep Convolutional Networks for Large-Scale Image Recognition*. https://doi.org/10.48550/ARXIV.1409.1556

Sun, L., Kamaliardakani, M., & Zhang, Y. (2016). Weighted Neighborhood Pixels Segmentation Method for Automated Detection of Cracks on Pavement Surface Images. *Journal of Computing in Civil Engineering*, *30*(2), 04015021. https://doi.org/10.1061/(ASCE)CP.1943-5487.0000488

Szegedy, C., Liu, W., Jia, Y., Sermanet, P., Reed, S., Anguelov, D., Erhan, D., Vanhoucke, V., & Rabinovich, A. (2014). *Going Deeper with Convolutions*. https://doi.org/10.48550/ARXIV.1409.4842

Tan, M., & Le, Q. V. (2019). *EfficientNet: Rethinking Model Scaling for Convolutional Neural Networks*. https://doi.org/10.48550/ARXIV.1905.11946

Tancik, M., Srinivasan, P. P., Mildenhall, B., Fridovich-Keil, S., Raghavan, N., Singhal, U., Ramamoorthi, R., Barron, J. T., & Ng, R. (2020). *Fourier Features Let Networks Learn High Frequency Functions in Low Dimensional Domains*. https://doi.org/10.48550/ARXIV.2006.10739

Tang, Y., Qian, Y., & Yang, E. (2022). Weakly supervised convolutional neural network for pavement crack segmentation. *Intelligent Transportation Infrastructure*, *1*, liac013. https://doi.org/10.1093/iti/liac013

Tello-Cifuentes, L., Marulanda, J., & Thomson, P. (2024). Detection and classification of pavement damages using wavelet scattering transform, fractal dimension by box-counting method and machine learning algorithms. *Road Materials and Pavement Design*, *25*(3), 566–584. https://doi.org/10.1080/14680629.2023.2219338

Tomar, N. K., Shergill, A., Rieders, B., Bagci, U., & Jha, D. (2022). *TransResU-Net: Transformer based ResU-Net for Real-Time Colonoscopy Polyp Segmentation* (arXiv:2206.08985). arXiv. http://arxiv.org/abs/2206.08985

Tsai, Y.-C., Kaul, V., & Mersereau, R. M. (2010). Critical Assessment of Pavement Distress Segmentation Methods. *Journal of Transportation Engineering*, *136*(1), 11–19. https://doi.org/10.1061/(ASCE)TE.1943-5436.0000051

Wang, S., & Tang, W. (2011). Pavement Crack Segmentation Algorithm Based on Local Optimal Threshold of Cracks Density Distribution. In D.-S. Huang, Y. Gan, V. Bevilacqua, & J. C. Figueroa (Eds.), *Advanced Intelligent Computing* (Vol. 6838, pp. 298–302). Springer Berlin Heidelberg. https://doi.org/10.1007/978-3-642-24728-6_40

Wang, W., & Su, C. (2021). Semi-supervised semantic segmentation network for surface crack detection. *Automation in Construction*, *128*, 103786. https://doi.org/10.1016/j.autcon.2021.103786

Wu, A. Y., Hong, T.-H., & Rosenfeld, A. (1982). Threshold Selection Using Quadtrees. *IEEE Transactions on Pattern Analysis and Machine Intelligence*, *PAMI-4*(1), 90–94. https://doi.org/10.1109/TPAMI.1982.4767203

Xu, Y., Lam, H.-K., & Jia, G. (2021). MANet: A two-stage deep learning method for classification of COVID-19 from Chest X-ray images. *Neurocomputing*, *443*, 96–105. https://doi.org/10.1016/j.neucom.2021.03.034

Yang, F., Zhang, L., Yu, S., Prokhorov, D., Mei, X., & Ling, H. (2019). *Feature Pyramid and Hierarchical Boosting Network for Pavement Crack Detection* (arXiv:1901.06340). arXiv. http://arxiv.org/abs/1901.06340

Yeung, M., Sala, E., Schönlieb, C.-B., & Rundo, L. (2022). Unified Focal loss: Generalising Dice and cross entropy-based losses to handle class imbalanced medical image segmentation. *Computerized Medical Imaging and Graphics*, *95*, 102026. https://doi.org/10.1016/j.compmedimag.2021.102026

Zhang, C., Nateghinia, E., Miranda-Moreno, L. F., & Sun, L. (2022). Pavement distress detection using convolutional neural network (CNN): A case study in Montreal, Canada.

*International Journal of Transportation Science and Technology*, *11*(2), 298–309. https://doi.org/10.1016/j.ijtst.2021.04.008

Zhang, H., Qian, Z., Tan, Y., Xie, Y., & Li, M. (2022). Investigation of pavement crack detection based on deep learning method using weakly supervised instance segmentation framework. *Construction and Building Materials*, *358*, 129117. https://doi.org/10.1016/j.conbuildmat.2022.129117

Zhang, Y., Gao, X., & Zhang, H. (2023). Deep Learning-Based Semantic Segmentation Methods for Pavement Cracks. *Information*, *14*(3), Article 3. https://doi.org/10.3390/info14030182

Zhao, H., Shi, J., Qi, X., Wang, X., & Jia, J. (2016). *Pyramid Scene Parsing Network*. https://doi.org/10.48550/ARXIV.1612.01105

Zhou, Z., Siddiquee, M. M. R., Tajbakhsh, N., & Liang, J. (2018). UNet++: A Nested U-Net Architecture for Medical Image Segmentation. *Deep Learning in Medical Image Analysis and Multimodal Learning for Clinical Decision Support : 4th International Workshop, DLMIA 2018, and 8th International Workshop, ML-CDS 2018, Held in Conjunction with MICCAI 2018, Granada, Spain, S*, *11045*, 3–11. https://doi.org/10.1007/978-3-030-00889-5_1

Zou, K. H., Warfield, S. K., Bharatha, A., Tempany, C. M. C., Kaus, M. R., Haker, S. J., Wells, W. M., Jolesz, F. A., & Kikinis, R. (2004). Statistical validation of image segmentation quality based on a spatial overlap index1. *Academic Radiology*, *11*(2), 178–189. https://doi.org/10.1016/S1076-6332(03)00671-8

Zuo, Y., Wang, G., & Zuo, C. (2008). A Novel Image Segmentation Method of Pavement Surface Cracks Based on Fractal Theory. *2008 International Conference on Computational Intelligence and Security*, 485–488. https://doi.org/10.1109/CIS.2008.206